\documentclass[10pt]{article} 
\usepackage[preprint]{tmlr}

\usepackage{hyperref}
\usepackage{url}

\usepackage{graphicx}
\usepackage{amsmath,amsfonts}
\usepackage{algorithmic}
\usepackage{algorithm}
\usepackage{array}
\usepackage{verbatim}
\usepackage{graphicx}
\usepackage{diagbox}
\usepackage{tcolorbox}
\usepackage{multirow}
\usepackage{multicol}
\usepackage{booktabs}
\usepackage{makecell}
\usepackage{siunitx}

\newcommand{\pb}[1]{\vspace{0.5ex}\noindent{\bf \em #1}\hspace*{.3em}}

\title{From Similarity to Structure: Training-free LLM Context Compression with Hybrid Graph Priors}

\author{\name Yitian Zhou,
      \name Chaoning Zhang \thanks{Corresponding author, email: chaoningzhang1990@gmail.com},
      \name Jiaquan Zhang,
      \name Zhenzhen Huang \\
      \addr School of Computer Science and Engineering \\
      University of Electronic Science and Technology of China
      \AND
      \name Jinyu Guo \\
      \addr School of Information and Software Engineering, 
      University of Electronic Science and Technology of China
      \AND
      \name Sung-Ho Bae \\
      \addr Department of Computer Science and Engineering, 
      Kyung Hee University
      \AND
      \name Lik-Hang Lee \\
      \addr Department of Industrial and Systems Engineering, 
      The Hong Kong Polytechnic University
      \AND
      \name Caiyan Qin \\
      \addr School of Robotics and Advanced Manufacture, 
      Harbin Institute of Technology, Shenzhen
      \AND
      \name Yang Yang \\
      \addr School of Computer Science and Engineering, 
      University of Electronic Science and Technology of China
      }

\def\month{MM}  
\def\year{YYYY} 
\def\openreview{\url{https://openreview.net/forum?id=XXXX}} 

\begin{document}

\maketitle

\begin{abstract}
Long-context large language models remain computationally expensive to run and often fail to reliably process very long inputs, which makes context compression an important component of many systems. Existing compression approaches typically rely on trained compressors, dense retrieval-style selection, or heuristic trimming, and they often struggle to jointly preserve task relevance, topic coverage, and cross-sentence coherence under a strict token budget. To address this, we propose a training-free and model-agnostic compression framework that selects a compact set of sentences guided by structural graph priors. Our method constructs a sparse hybrid sentence graph that combines mutual k-NN semantic edges with short-range sequential edges, extracts a topic skeleton via clustering, and ranks sentences using an interpretable score that integrates task relevance, cluster representativeness, bridge centrality, and a cycle coverage cue. A budgeted greedy selection with redundancy suppression then produces a readable compressed context in original order. Experimental results on four datasets show that our approach is competitive with strong extractive and abstractive baselines, demonstrating larger gains on long-document benchmarks.
\end{abstract}

\section{Introduction}

Large language models (LLMs) have achieved remarkable success in natural language processing tasks such as question answering and summarization, but their performance is constrained by the context window, i.e., the maximum tokens processed in one forward pass. Although existing LLMs support increasingly large windows, such as 128K tokens in GPT-4o and 400K tokens in GPT-5, processing extremely long inputs remains costly. Standard Transformer attention scales quadratically with sequence length \cite{vaswani2017attention}, motivating efficiency-oriented long-context architectures like sparse-attention \cite{beltagy2020longformer}. More importantly, longer windows do not guarantee reliable use of long context, as LLM performance can drop when relevant evidence is placed in the middle of a long context \cite{liu2024lost}. Recent long-context evaluations also report sizable degradations as context length grows beyond nominal claims \cite{hsieh2024ruler}. As a result, real systems still often rely on truncation, segmentation, or multi-stage pipelines, which can break cross-segment coherence and remove critical evidence, thereby reducing downstream task performance.

Recent research addressing LLM with long inputs has shifted to how to keep fewer but better tokens, apart from how to fit more tokens, compressing long-context while keeping semantic efficiency. 
Research has revisited the classical connection between prediction and compression. Traditional lossless text compression based on dictionaries and statistical models is well established and widely documented \cite{bell1989modeling,sayood2017introduction}. Recent work re-centers this topic around modern autoregressive language models, showing that strong next-token predictors can be turned into strong lossless compressors when paired with arithmetic coding \cite{deletang2024language}, and presenting concrete LM-based compression pipelines that achieve competitive compression rates \cite{valmeekam2023llmzip}. This direction clarifies why LMs can compress in principle, but it also highlights that lossless compression is token-level and does not explicitly aim to preserve task-specific evidence, topic balance, or discourse structure that users need for reasoning and interpretability \cite{zhang2026learning}.
Recent studies further explore compression from a model-efficiency perspective, such as spike-driven lightweight architectures and brain-inspired sequential memory mechanisms, which suggest that structured information retention, rather than raw token capacity, is critical for scalable reasoning \cite{11152565,10924795}.

In parallel, context compression has rapidly evolved from simple heuristic trimming to methods that explicitly optimize for downstream utility under a budget. A representative trend is to treat compression as a learned selection or transformation problem, where a smaller model is trained to remove redundant parts while keeping what the LLM needs.
For example, LLMLingua-2 casts prompt compression as token classification derived via data distillation \cite{pan2024llmlingua}, while QUITO leverages query-guided attention signals to filter context for long-context reasoning \cite{wang2024quito}. More recent efforts go further toward implicit or representation-level compression, aiming to pack information into fewer soft units or learned summaries and then recover task performance after compression \cite{dai2025pretraining,tang2025gmsa}.
Related ideas have also been explored in retrieval-augmented generation and dialogue systems, where selective context utilization and redundancy reduction are critical for efficiency and robustness \cite{guo2025hash,ou2025accelerating}.
These advances improve effectiveness, but they often add training stages, specialized compressors, or less transparent intermediate representations, which can be a barrier when the goal is a plug-and-play and easily explainable preprocessor \cite{liskavets2025prompt,kinger2025enhancing}.

For interpretability, graph-based methods are a natural choice because they explicitly model relationships between semantic units, allowing structural signals to guide selection. Classic unsupervised extractive summarization methods build a similarity graph and rank sentences by centrality  \cite{mihalcea2004textrank,erkan2004lexrank}. More recently, researchers have explored richer graph constructions for long documents that incorporate multiple edge types and structural cues, aiming to better capture long-range dependencies and document organization \cite{bugueno2025graphlss,onan2024ketgs}. However, as graph designs become more complex, they can also become denser and more redundant, and many modern graph pipelines rely on learned components to achieve robust generalization, which again moves away from the lightweight, training-free goal.

To overcome these issues, we aim for a context compressor that is simultaneously training-free, model-agnostic, and explainable, while still being sensitive to global structure. Our core hypothesis is that a sparse graph with carefully chosen priors can capture the minimum structure needed for robust compression without introducing a heavy learned graph encoder. Concretely, we build a hybrid sentence graph where semantic relations are represented by a mutual $k$-NN similarity graph and discourse flow is approximated by short-range sequential edges, so the graph reflects both ``what is about the same thing'' and ``what is connected in the narrative.'' We then extract a topic skeleton via lightweight clustering and score each sentence with task relevance and three interpretable structural priors: cluster representativeness within its topic (to encourage coverage), bridge importance via betweenness centrality (to protect cross-topic connectors), and cycle coverage (to reduce the chance of cutting local reasoning loops). Betweenness centrality provides a standard way to quantify bridge nodes in a network \cite{freeman1977betweenness}, and cycle-basis structure offers a tractable signal for closed connectivity patterns \cite{paton1969cycles}. Under a strict token budget, we perform greedy selection with redundancy control in the spirit of relevance–novelty selection \cite{carbonell1998mmr}, and we output selected sentences in original order to preserve readability.
These factors preserve topic coverage and protect cross-segment connections and coherence in semantic chains. The final output is evaluated with respect to compression ratio, task performance, and structural preservation. 

The main contributions of this paper are summarized as follows:

(1) We propose a training-free and model-agnostic context compression framework that constructs a hybrid sentence graph by jointly modeling semantic similarity and physical order, enabling structure-aware extraction under token budgets.

(2) We develop an interpretable scoring-and-selection pipeline that combines task relevance, topic representativeness, and graph structural priors, with greedy budgeted selection and redundancy control to preserve coverage and coherence.

(3) We validate the method with extensive budgeted compression experiments and ablations, measuring downstream task performance and the impact of each component.

\section{Related Work}

In this section, we review prior work along three related lines, including long-document modeling, extractive sentence selection, and LLM-oriented context compression.

\subsection{Long-Document Summarization and Long-Context Modeling}

A common approach for long documents is to extend the model's usable context \cite{zhang2026text}. Since standard self-attention has quadratic cost in the input length \cite{vaswani2017attention}, sparse-attention Transformers such as Longformer \cite{beltagy2020longformer} and BIGBIRD \cite{zaheer2020big} reduce attention complexity and enable longer contexts. In summarization, pretrained seq2seq models like BART \cite{lewis2020bart} and PEGASUS \cite{zhang2020pegasus} generate high-quality summaries, but still face efficiency and stability issues with extremely long inputs, leading to multi-stage pipeline designs. For example, DANCER \cite{gidiotis2020dancer} uses discourse structure and decomposition to handle long documents more effectively. These approaches improve scalability, but they may weaken global structure modeling and break cross-segment dependencies, which motivates budgeted compression methods that better preserve document-level connections \cite{zhang2026learning,10924795}.

\subsection{Extractive Sentence Selection and Graph-Based Ranking}

Extractive summarization remains well-suited for budgeted context compression because it preserves original wording and allows direct auditing of retained evidence. Classic unsupervised graph-based methods such as TextRank \cite{mihalcea2004textrank} and LexRank \cite{erkan2004lexrank} build sentence similarity graphs and rank sentences by centrality. In practice, positional heuristics like LEAD-3 are also strong baselines, especially for news-style documents, and are widely used in extractive summarization evaluations \cite{nallapati2016abstractive}. Neural extractive methods improve selection quality by learning sentence importance from contextual encoders and better training objectives.
BERTSum fine-tunes BERT-style encoders for sentence extraction \cite{liu2019bertsum}, MatchSum reframes extraction as semantic matching between candidate subsets and the source document \cite{zhong2020matchsum}, and MemSum explicitly models extraction history in a multi-step decision process to reduce redundancy and better cover long documents \cite{gu2022memsum}. Recent training-free approaches also explore combining multiple signals for sentence salience, such as RankSum, which performs rank fusion over topic, semantic, keyword, and positional features \cite{joshi2022ranksum}. These works are closely related to our setting.
Beyond summarization, similar selection mechanisms have been studied in dialogue state tracking \cite{guo2022beyond}.
Still, most similarity-only selection methods do not explicitly protect bridge sentences or discourse loops, and many neural approaches rely on supervised training. At the same time, we target training-free compression with explicit structure cues.

\subsection{LLM Context and Prompt Compression}
Beyond summarization as an end task, recent work studies compression as a preprocessing module for LLMs, aiming to reduce tokens while keeping downstream performance \cite{wang2026stream,zhang2026lightweight}. LLMLingua-2 learns prompt compression via token classification using data distillation \cite{pan2024llmlingua}, and QUITO uses query-guided attention signals to filter context for long-context reasoning under budget constraints \cite{wang2024quito}. Other methods guide LLM summarization with explicit intermediate signals rather than direct end-to-end rewriting. SigExt injects extracted salient keyphrases into prompts to improve content control in prompt-based abstractive summarization \cite{xu2024sigext}, while highlight-guided self-planning uses sentence-level highlights for faithful generation \cite{du2025higen}. 
While effective, many of these methods introduce additional training stages, use token-level operations that are less transparent at the discourse level, or produce intermediate representations that are not straightforward to audit for structural integrity \cite{wang2025think}. Recent multimodal LLM studies further highlight that compression and reasoning should be jointly optimized, as reducing redundant representations while preserving reasoning chains is key to maintaining performance under tight budgets \cite{zheng2026LLaVA,li2026experience}.
Our work is inspired by these studies to propose sentence-level, training-free compression under a strict token budget.

\begin{figure*}[t]
\includegraphics[width=\linewidth]{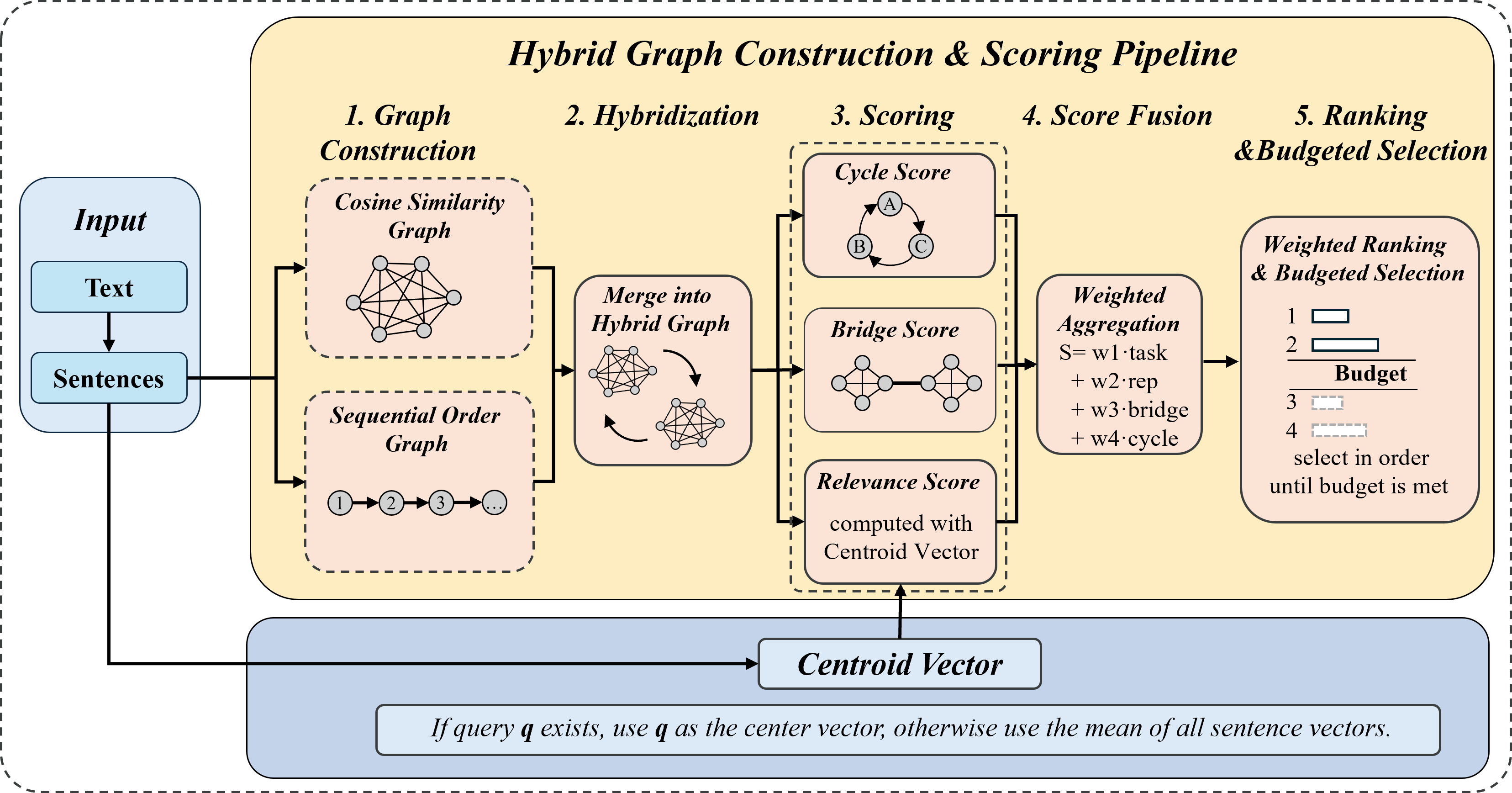}
\caption{Overview of the proposed structure-aware context compression framework. }
\label{fig1}
\end{figure*}

\section{Method}

In this section, we describe the overall structure and key modules of our proposed training-free and model-agnostic context compression framework. Our goal is to compress a long document into a compact, sentence-level subset that fits a token budget while remaining useful for downstream LLM tasks. To make sentence selection explicitly \emph{structure-aware}, our method constructs a sparse hybrid graph over sentences, extracts a topic skeleton via clustering, and then scores each sentence using interpretable priors that reflect relevance, coverage, and connectivity. Sentences are selected under a strict budget using a greedy procedure with redundancy suppression, and the final compressed context is formed by restoring the original sentence order. Fig.~\ref{fig1} illustrates the overall workflow.

\subsection{Problem Formulation}
Given a document represented as a sentence sequence $\mathcal{D}=\{s_1,\dots,s_N\}$ and an optional task description or query $q$, we seek a subset $\mathcal{S}\subset \mathcal{D}$ that satisfies a token budget constraint. Let $\mathrm{Tok}(\cdot)$ denote the token count under the target tokenizer. The constraint is $\mathrm{Tok}(\mathcal{S}) \le B$, where $B$ can also be specified implicitly via a compression ratio $\rho$ such that $B=\rho\cdot \mathrm{Tok}(\mathcal{D})$. Under this constraint, the selection should jointly promote (i) task relevance, (ii) topic coverage, and (iii) structural coherence, i.e., avoiding the removal of sentences that act as bridges in argument chains or as anchors in local discourse loops.



\subsection{Sentence Embedding}

We first split the input into sentences and encode each sentence into a dense vector representation using a pretrained embedding model:
\begin{equation}\label{eq:emb1}
\mathbf{e}_i = \mathrm{Embed}(s_i), \quad \mathbf{e}_i \in \mathbb{R}^d.
\end{equation}
All embeddings are $L_2$-normalized to stabilize cosine similarity:
\begin{equation}\label{eq:emb2}
\mathbf{e}_i \leftarrow \frac{\mathbf{e}_i}{\| \mathbf{e}_i \| _2}.
\end{equation}
In implementation, we cap each sentence at 512 tokens when computing embeddings to bound computation, which rarely truncates natural sentences in typical corpora, and avoids pathological cases such as noisy line breaks or merged paragraphs. These embeddings are used to build a semantic graph that captures both local and global relations among sentences.

\subsection{Hybrid Sentence Graph Construction}
We build a sentence-level graph that combines two complementary signals: semantic proximity and physical adjacency. Semantic edges capture topical similarity, while sequential edges preserve local narrative and discourse continuity that may not be reflected by embedding neighbors.

\pb{Semantic similarity graph.}
We construct a mutual $k$-nearest neighbor ($k$-NN) graph $G_{\text{sem}}=(V, E_{\text{sem}})$ where each node corresponds to a sentence. For each node $i$, we find its top-$k$ neighbors by cosine similarity and retain an edge $(i,j)$ only if $i$ is in the top-$k$ list of $j$ and vice versa. This mutual constraint reduces one-way noisy connections and helps keep the graph sparse. We use approximate nearest neighbor search (HNSW) for scalability. Larger $k$ values are generally suited for longer texts to ensure connectivity, while smaller $k$ values keep the graph sparse. Unless otherwise stated, we use $k=8$, which empirically provides sufficient connectivity for long documents while keeping the number of edges near linear in $N$. 
The weight of each edge is defined as:
\begin{equation}\label{eq:wsem}
w^{\text{sem}}_{ij} = \cos(\mathbf{e}_i,\mathbf{e}_j),
\end{equation}
where $e_i$and $e_j$ are normalized embeddings. Note that cosine similarity is already bounded in $[-1,1]$, while with normalized embeddings in practice it typically concentrates in $[0,1]$ for semantically related sentences.

\pb{Sequential edges.}
To encode local discourse flow, we add sequential edges between sentences within a small window:
\begin{equation}\label{eq:eseq}
E_{\text{seq}}=\{(i,j)\mid |i-j|\le \Delta\},
\end{equation}
where we set $\Delta=1$ by default to capture immediate neighbors. And the sequential edge weight decays with distance:
\begin{equation}\label{eq:wseq}
w^{\text{seq}}_{ij} = \exp(-|i-j|).
\end{equation}

\pb{Graph fusion.} 
The hybrid graph is then denoted as $G=(V,E)$ with $E=E_{\text{sem}}\cup E_{\text{seq}}$, and edges are weighted:
\begin{equation}\label{eq:w}
\lambda_{ij}=\alpha \cdot w^{\text{sem}}_{ij} + \beta \cdot w^{\text{seq}}_{ij}, 
\end{equation}
where $\alpha+\beta=1,$ and $\alpha,\beta\ge 0.$
Intuitively, $\alpha$ controls how much we trust semantic neighborhoods, while $\beta$ ensures we do not completely discard local narrative structure. Smaller $\beta$ places more emphasis on semantic neighborhoods, while larger $\beta$ strengthens local discourse continuity. We keep the graph sparse by design, which is important both for efficiency and for avoiding redundancy introduced by dense connectivity.

\subsection{Topic Skeleton via Graph-based Clustering}
Once the semantic graph has been constructed, we estimate the latent topical structure of the document by clustering sentence embeddings to encourage coverage. Specifically, we apply \emph{MiniBatch K-Means}, which is computationally efficient for large collections of sentence embeddings while maintaining clustering quality comparable to standard K-Means. The number of clusters is determined by the square-root heuristic:
\begin{equation}\label{eq:k}
K = \mathrm{round}(\sqrt{N}),
\end{equation}
where $N$ denotes the number of sentences. A larger $K$ risks producing fragmented clusters that capture noise rather than salient themes, while a smaller $K$ may merge semantically distinct content and obscure the diversity of topics. Each sentence $i$ is assigned a cluster label $c(i)$ with centroid vector $\mathbf{c}_{c(i)}$. These centroids serve as semantic anchors representing the thematic skeleton of the text. 
And topic representativeness is defined with cosine similarity:
\begin{equation}\label{eq:rep}
S_{\text{rep}}(i)=\cos(\mathbf{e}_i,\mathbf{c}_{c(i)}).
\end{equation}
This term favors sentences that are central within a topic region, which helps avoid selecting only edging or overly specific sentences, even when they are locally relevant.

\subsection{Structure-Aware Sentence Scoring}
After clustering, each sentence is assigned a composite score that integrates task relevance, topic representativeness, and two other graph-structural priors that reflect global connectivity and local loop structure.

\pb{Task relevance.}
If a query $q$ is provided, we embed it as $\mathbf{e}_q$ and define the task relevance score $S_{\text{task}}$ as:
\begin{equation}\label{eq:task1}
S_{\text{task}}(i)=\cos(\mathbf{e}_i,\mathbf{e}_q).
\end{equation}
If no query is available, we use the document centroid $\bar{\mathbf{e}}=\frac{1}{N} \sum_{i=1}^{N}\mathbf{e}_i$ as a proxy for global salience:
\begin{equation}\label{eq:task2}
S_{\text{task}}(i)=\cos(\mathbf{e}_i,\bar{\mathbf{e}}).
\end{equation}

\pb{Bridge centrality.}
To protect cross-topic and cross-section links, we compute the betweenness ($\mathrm{Btw}$) centrality on the hybrid graph:
\begin{equation}\label{eq:btw}
S_{\text{bridge}}(i)=\mathrm{Btw}(i;G),
\end{equation}
and then normalize it to $[0,1]$. In practice, exact betweenness can be expensive for large graphs; we therefore use an approximate variant with sampling, where the number of sampled sources is set proportional to $\sqrt{N}$ to balance stability and cost:
\begin{equation}\label{eq:bridge}
S_{\text{bridge}}(i)=\mathrm{ApprBtw}(i;G,\mathrm{samples}=\lceil \sqrt{N}\rceil).
\end{equation}
This term promotes sentences that lie on many shortest paths, which often correspond to rhetorical transitions, definitions that connect sections, or key evidence linking claims to conclusions.

\pb{Cycle coverage cue.}
Similarity graphs of coherent text often contain small cycles that reflect local mutual reinforcement, such as claim--evidence pairs that connect through shared concepts, or recurring references. To reduce the risk of breaking such local closure patterns, we identify a limited set of small cycles through a cycle basis or bounded cycle enumeration, and mark whether a sentence participates in any of them:
\begin{equation}\label{eq:cycle}
S_{\text{cycle}}(i)=
\begin{cases}
1, & i \in \mathcal{C}, \\
0, & \text{otherwise},
\end{cases}
\end{equation}
where $\mathcal{C}$ denotes the set of nodes covered by the enumerated cycles. For efficiency, we cap the enumeration to a fixed number (e.g., 200 cycles) and treat this term as a binary cue rather than a fine-grained score.

\pb{Composite score.}
With the four scores computed in Eq. \eqref{eq:rep}-\eqref{eq:cycle}, the final score is a weighted sum:
\begin{equation}\label{eq:score}
\begin{split}
S(i)=&\lambda_{\text{task}} S_{\text{task}}(i) + \lambda_{\text{rep}} S_{\text{rep}}(i)+\\&\lambda_{\text{bridge}} S_{\text{bridge}}(i) + \lambda_{\text{cycle}} S_{\text{cycle}}(i),
\end{split}
\end{equation}
where $\lambda_{\text{task}},\lambda_{\text{rep}},\lambda_{\text{bridge}},\lambda_{\text{cycle}}$ are interpretable hyperparameters. Unless otherwise specified, we use a fixed setting that prioritizes relevance and coverage while still rewarding connectivity:
\begin{equation}
(\lambda_{\text{task}},\lambda_{\text{rep}},\lambda_{\text{bridge}},\lambda_{\text{cycle}})=(0.45,0.3,0.2,0.05).
\end{equation}
We test small perturbations of these weights and find the relative performance trends remain stable, suggesting robustness to weight variation. 

\subsection{Budgeted Selection with Redundancy Suppression}
Given sentence scores, we perform greedy selection under a token budget. Sentences are sorted by $S(i)$ in descending order, and we add a candidate sentence $s_i$ to $\mathcal{S}$ if it satisfies two constraints: budget feasibility and non-redundancy.

\pb{Budget feasibility.}
Let $B$ be the token budget. We add $s_i$ only if:
\begin{equation}
\mathrm{Tok}(\mathcal{S}\cup \{s_i\}) \le B.
\end{equation}
By default, we use a compression ratio of $\rho=0.30$ (i.e., $B=0.30\cdot \mathrm{Tok}(\mathcal{D})$), and we also evaluate other budgets to reflect different constraints.

\pb{Redundancy suppression.}
To avoid selecting many near-duplicate sentences, we apply a cosine-similarity non-maximum suppression (NMS) rule. A candidate $s_i$ is discarded if there exists a selected sentence $s_j\in \mathcal{S}$ such that:
\begin{equation}
\cos(\mathbf{e}_i,\mathbf{e}_j)\ge \tau.
\end{equation}
We use $\tau=0.92$ unless otherwise stated, which filters paraphrases and highly overlapping sentences while retaining complementary evidence.

The selected set $\mathcal{S}$ is finally sorted by original sentence index to form the compressed text, ensuring readability and preserving the document's original discourse flow.

\subsection{Output Evaluation}
The output of our method is a sentence subset that is directly consumable by any downstream LLM without additional training. We evaluate effectiveness along three dimensions:
(1) compression ratio $\mathrm{CR}=\mathrm{Tok}(\mathcal{S})\,/\,\mathrm{Tok}(\mathcal{D})$ and budget adherence; 
(2) downstream task effectiveness under different token budgets using task-appropriate metrics;
(3) structural preservation using lightweight proxies aligned with our design, including topic coverage rate measured by the fraction of clusters represented in $\mathcal{S}$, and graph-structure retention, such as the fraction of selected sentences that are bridge nodes or cycle-covered nodes in the original graph. 

\begin{table*}[ht]
\centering
\small
\setlength{\tabcolsep}{3pt}
\resizebox{\linewidth}{!}{
\begin{tabular}{l|ccccc|ccccc|ccccc|ccccc}
\toprule
\multirow{2}{*}{\textbf{Method}} &
\multicolumn{5}{c|}{\textbf{CNN/DailyMail}} &
\multicolumn{5}{c|}{\textbf{GovReport}} &
\multicolumn{5}{c|}{\textbf{arXiv}} &
\multicolumn{5}{c}{\textbf{PubMed}} \\
\cmidrule{2-21}
& \textbf{R-1} & \textbf{R-2} & \textbf{R-L} & \textbf{BS} & \textbf{FE}
& \textbf{R-1} & \textbf{R-2} & \textbf{R-L} & \textbf{BS} & \textbf{FE}
& \textbf{R-1} & \textbf{R-2} & \textbf{R-L} & \textbf{BS} & \textbf{FE}
& \textbf{R-1} & \textbf{R-2} & \textbf{R-L} & \textbf{BS} & \textbf{FE} \\
\midrule
TextRank & 33.10 & 12.20 & 29.70 & 0.73 & 0.58 & 53.19 & 23.12 & 49.86 & 0.91 & 0.81 & 33.10 & 8.80 & 30.05 & 0.73 & 0.58 & 38.66 & 15.87 & 34.53 & 0.80 & 0.70 \\
LEAD-3 & 39.94 & 17.46 & 36.06 & 0.83 & 0.68 & 50.94 & 19.53 & 48.45 & 0.89 & 0.79 & 25.53 & 5.98 & 15.22 & 0.65 & 0.50 & 26.38 & 8.73 & 16.60 & 0.68 & 0.55 \\
BERTSum & 41.63 & 19.44 & 40.13 & 0.87 & 0.77 & 48.83 & 20.46 & 45.15 & 0.88 & 0.78 & 47.10 & 18.20 & 32.40 & 0.80 & 0.70 & 49.10 & \textbf{24.30} & 36.40 & \textbf{0.85} & 0.75 \\
MatchSum & 44.41 & 20.86 & 40.55 & 0.88 & 0.78 & 50.23 & 21.34 & 46.05 & 0.89 & 0.79 & 45.06 & 17.83 & 40.21 & 0.81 & 0.73 & 41.21 & 14.91 & 36.75 & 0.82 & 0.72 \\
MemSum & 43.22 & 20.36 & 40.18 & 0.87 & 0.77 & 49.14 & 22.92 & 44.33 & 0.88 & 0.78 & 48.23 & 20.17 & 42.31 & 0.83 & 0.75 & 49.14 & 22.92 & 44.33 & \textbf{0.85} & 0.75 \\
BART & 44.16 & 21.28 & 40.09 & 0.86 & 0.77 & 52.24 & 22.09 & 49.99 & 0.90 & 0.80 & 43.84 & 16.55 & 39.86 & 0.79 & 0.70 & 44.61 & 19.37 & 41.01 & 0.84 & 0.74 \\
PEGASUS & 44.17 & 21.47 & 41.11 & \textbf{0.89} & 0.79 & 54.29 & 20.80 & 51.35 & 0.91 & 0.81 & 43.27 & 19.70 & 34.79 & 0.80 & 0.72 & 44.70 & 17.27 & 39.10 & 0.83 & 0.72 \\
BIGBIRD & 43.83 & 21.11 & 40.74 & 0.87 & 0.78 & 60.64 & 24.81 & 50.01 & 0.92 & 0.82 & 46.63 & 19.02 & 41.77 & 0.82 & 0.74 & 46.32 & 20.65 & 42.33 & \textbf{0.85} & 0.75 \\
DANCER & 43.57 & 20.86 & 40.52 & 0.88 & 0.78 & 57.89 & 23.16 & 49.67 & 0.91 & 0.81 & 45.01 & 17.60 & 40.56 & 0.81 & 0.73 & 46.34 & 19.97 & 42.42 & \textbf{0.85} & 0.75 \\
RankSum & \textbf{44.89} & \textbf{24.25} & \textbf{41.72} & 0.88 & 0.78 & 56.74 & 22.68 & 48.92 & 0.90 & 0.80 & 46.15 & 19.48 & 41.93 & 0.82 & 0.74 & 47.86 & 22.14 & 41.78 & 0.84 & 0.73 \\
HiGen & 44.32 & 21.35 & 41.02 & 0.88 & 0.77 & 53.64 & 21.86 & 47.18 & 0.91 & 0.83 & 45.72 & 19.21 & 42.08 & 0.83 & 0.75 & 48.21 & 22.63 & 43.05 & \textbf{0.85} & 0.74 \\
SigExt & 42.39 & 19.85 & 40.21 & 0.87 & 0.75 & 55.92 & 22.13 & 49.08 & 0.91 & 0.80 & 43.62 & 18.97 & 24.29 & 0.82 & 0.74 & 47.92 & 22.31 & 42.68 & \textbf{0.85} & 0.73 \\
Ours & 44.69 & 21.72 & 41.37 & \textbf{0.89} & \textbf{0.80} & \textbf{61.39} & \textbf{25.19} & \textbf{52.01} & \textbf{0.93} & \textbf{0.84} & \textbf{48.67} & \textbf{21.98} & \textbf{45.57} & \textbf{0.85} & \textbf{0.78} & \textbf{49.48} & 23.41 & \textbf{44.71} & \textbf{0.85} & \textbf{0.76} \\
\bottomrule
\end{tabular}}
\caption{Performance comparison between our proposed method and representative baselines on CNN/DailyMail, GovReport, arXiv, and PubMed. Best scores are in bold.}
\label{tab:main_all}
\end{table*}

\begin{table*}[ht]
\centering
\small
\setlength{\tabcolsep}{3pt}
\resizebox{\linewidth}{!}{
\begin{tabular}{l|ccccc|ccccc|ccccc|ccccc}
\toprule
\multirow{2}{*}{\textbf{Method}} &
\multicolumn{5}{c|}{\textbf{CNN/DailyMail}} &
\multicolumn{5}{c|}{\textbf{GovReport}} &
\multicolumn{5}{c|}{\textbf{arXiv}} &
\multicolumn{5}{c}{\textbf{PubMed}} \\
\cmidrule{2-21}
& \textbf{R-1} & \textbf{R-2} & \textbf{R-L} & \textbf{BS} & \textbf{FE}
& \textbf{R-1} & \textbf{R-2} & \textbf{R-L} & \textbf{BS} & \textbf{FE}
& \textbf{R-1} & \textbf{R-2} & \textbf{R-L} & \textbf{BS} & \textbf{FE}
& \textbf{R-1} & \textbf{R-2} & \textbf{R-L} & \textbf{BS} & \textbf{FE} \\
\midrule
Ours (Full) & 44.69 & 21.72 & 41.37 & 0.89 & 0.80 & 61.39 & 25.19 & 52.01 & 0.93 & 0.84 & 48.67 & 21.98 & 45.57 & 0.85 & 0.78 & 49.48 & 23.41 & 44.71 & 0.85 & 0.76 \\
w/o Seq & 44.53 & 21.54 & 41.05 & 0.89 & 0.79 & 58.92 & 23.81 & 49.28 & 0.92 & 0.82 & 46.92 & 20.42 & 42.68 & 0.83 & 0.76 & 48.52 & 22.58 & 42.96 & 0.85 & 0.75 \\
w/o Rep & 44.21 & 21.08 & 40.92 & 0.88 & 0.79 & 59.76 & 24.02 & 50.12 & 0.92 & 0.82 & 47.28 & 20.73 & 43.61 & 0.84 & 0.76 & 48.68 & 22.74 & 43.37 & 0.85 & 0.75 \\
w/o Bridge & 44.35 & 21.37 & 40.88 & 0.88 & 0.79 & 59.41 & 24.15 & 49.36 & 0.92 & 0.82 & 46.84 & 20.58 & 42.74 & 0.83 & 0.76 & 48.47 & 22.61 & 43.05 & 0.85 & 0.75 \\
w/o Cycle & 44.58 & 21.63 & 41.21 & 0.89 & 0.80 & 60.72 & 24.93 & 51.17 & 0.93 & 0.83 & 48.05 & 21.47 & 44.53 & 0.84 & 0.77 & 49.11 & 23.09 & 44.07 & 0.85 & 0.76 \\
w/o NMS & 44.05 & 20.92 & 40.74 & 0.88 & 0.79 & 59.58 & 23.62 & 50.03 & 0.92 & 0.82 & 47.35 & 20.55 & 43.95 & 0.84 & 0.78 & 48.62 & 22.46 & 43.41 & 0.85 & 0.76 \\
\bottomrule
\end{tabular}}
\caption{Ablation results of different components of our framework on four benchmarks.}
\label{tab:ablation_all}
\end{table*}

\begin{table*}[ht]
\centering
\small
\setlength{\tabcolsep}{3pt}
\resizebox{\linewidth}{!}{
\begin{tabular}{cc|ccccc|ccccc|ccccc|ccccc}
\toprule
\multirow{2}{*}{$\boldsymbol{\beta}$} & \multirow{2}{*}{$\boldsymbol{\alpha}$} &
\multicolumn{5}{c|}{\textbf{CNN/DailyMail}} &
\multicolumn{5}{c|}{\textbf{GovReport}} &
\multicolumn{5}{c|}{\textbf{arXiv}} &
\multicolumn{5}{c}{\textbf{PubMed}} \\
\cmidrule{3-22}
& & \textbf{R-1} & \textbf{R-2} & \textbf{R-L} & \textbf{BS} & \textbf{FE}
& \textbf{R-1} & \textbf{R-2} & \textbf{R-L} & \textbf{BS} & \textbf{FE}
& \textbf{R-1} & \textbf{R-2} & \textbf{R-L} & \textbf{BS} & \textbf{FE}
& \textbf{R-1} & \textbf{R-2} & \textbf{R-L} & \textbf{BS} & \textbf{FE} \\
\midrule
0.00 & 1.00 & 44.53 & 21.54 & 41.05 & 0.89 & 0.79 & 58.92 & 23.81 & 49.28 & 0.92 & 0.82 & 46.92 & 20.42 & 42.68 & 0.83 & 0.76 & 48.52 & 22.58 & 42.96 & 0.85 & 0.75 \\
0.25 & 0.75 & 44.62 & 21.66 & 41.22 & 0.89 & 0.80 & 60.10 & 24.42 & 50.63 & 0.92 & 0.83 & 47.85 & 21.02 & 44.01 & 0.84 & 0.77 & 48.96 & 22.97 & 43.81 & 0.85 & 0.75 \\
0.50 & 0.50 & 44.67 & 21.70 & 41.30 & 0.89 & 0.80 & 60.88 & 24.86 & 51.34 & 0.93 & 0.83 & 48.31 & 21.54 & 44.83 & 0.84 & 0.77 & 49.21 & 23.19 & 44.33 & 0.85 & 0.76 \\
0.75 & 0.25 & \textbf{44.69} & \textbf{21.72} & \textbf{41.37} & \textbf{0.89} & \textbf{0.80} & \textbf{61.39} & \textbf{25.19} & \textbf{52.01} & \textbf{0.93} & \textbf{0.84} & \textbf{48.67} & \textbf{21.98} & \textbf{45.57} & \textbf{0.85} & \textbf{0.78} & \textbf{49.48} & \textbf{23.41} & \textbf{44.71} & \textbf{0.85} & \textbf{0.76} \\
1.00 & 0.00 & 44.30 & 21.28 & 40.88 & 0.88 & 0.79 & 59.95 & 24.10 & 50.08 & 0.92 & 0.82 & 47.42 & 20.86 & 43.72 & 0.83 & 0.76 & 48.71 & 22.84 & 43.59 & 0.85 & 0.70 \\
\bottomrule
\end{tabular}}
\caption{Effect of the fusion weight $\beta$ (with $\alpha=1-\beta$) on hybrid-graph construction. }
\label{tab:beta_scan}
\end{table*}

\begin{figure*}[t]
\includegraphics[width=\linewidth]{Fig/compress_cost.png}
\caption{Performance over token budgets on four datasets. We report the most discriminative metric per dataset (QAFactEval for CNN/DailyMail and GovReport, ROUGE-L for arXiv and PubMed).}
\label{fig:cost}
\end{figure*}

\section{Experiments}

We evaluate our hybrid-graph context compression method on four widely used summarization benchmarks and compare against representative extractive, abstractive, and long-context baselines under the same token budget settings.

\subsection{Experimental Setup}
\pb{Datasets.} 
We evaluate our method on four commonly-used summarization datasets that cover both short-to-medium news articles and long-form documents, including \textbf{CNN/DailyMail} \cite{hermann2015teaching}, \textbf{GovReport} \cite{huang2021efficient}, \textbf{arXiv} \cite{cohan-etal-2018-discourse}, and \textbf{PubMed} \cite{cohan-etal-2018-discourse}.

\pb{Baselines.} 
We compare against a diverse set of representative methods for summarization and compression, including \textbf{TextRank} \cite{mihalcea2004textrank}, \textbf{LEAD-3}, \textbf{BERTSum} \cite{liu2019bertsum}, \textbf{MatchSum} \cite{zhong2020matchsum}, \textbf{MemSum} \cite{gu2022memsum}, \textbf{BART} \cite{lewis2020bart}, \textbf{PEGASUS} \cite{zhang2020pegasus}, \textbf{BIGBIRD} \cite{zaheer2020big}, \textbf{DANCER} \cite{gidiotis2020dancer}, \textbf{RankSum} \cite{joshi2022ranksum}, \textbf{HiGen} \cite{du2025higen}, and \textbf{SigExt} \cite{xu2024sigext}. This baseline set allows us to compare against (i) similarity-based extraction, (ii) supervised extraction, (iii) abstractive generation, (iv) long-context modeling, and (v) recent LLM prompting and planning methods for context compression.

\pb{Ablation settings.}
We conduct ablation studies on five components: (i) \textbf{w/o Seq} sets $\beta=0$ ($\alpha=1$) to remove sequential edges and use only mutual $k$-NN semantic edges; (ii) \textbf{w/o Rep} sets $\lambda_{\text{rep}}=0$ to drop cluster representativeness (clustering is still performed for comparability); (iii) \textbf{w/o Bridge} sets $\lambda_{\text{bridge}}=0$ to remove the betweenness-based bridge reward; (iv) \textbf{w/o Cycle} sets $\lambda_{\text{cycle}}=0$ to remove the cycle-coverage bonus; and (v) \textbf{w/o NMS} disables the redundancy filter by removing the similarity threshold $\tau$ so sentences are selected purely by descending score until the budget $B$ is met.

\pb{Metrics.} 
We evaluate summarization quality with \textbf{ROUGE-1/2/L} \cite{lin2004rouge} and \textbf{BERTScore (BS)} \cite{zhang2019bertscore}, and assess factual consistency using \textbf{QAFactEval (FE)} \cite{fabbri2022qafacteval}.

Further details are presented in the Appendix.

\subsection{Performance Evaluation}

As shown in Table \ref{tab:main_all}, our method performs consistently well across domains and document lengths, with particularly strong gains on long-document benchmarks. On GovReport and arXiv, we achieve the best results on all metrics, suggesting that explicitly modeling topic coverage and cross-sentence structure is especially beneficial when salient evidence is distributed across distant sections. On PubMed, our method yields the best ROUGE-1/ROUGE-L and QAFactEval, indicating improved overall alignment and factual consistency. On CNN/DailyMail, while RankSum attains the highest ROUGE scores, our approach remains competitive and achieves the best QAFactEval and BERTScore, reflecting better faithfulness despite similar overlap-based quality.

\subsection{Ablation Studies}

Table~\ref{tab:ablation_all} shows that each component contributes positively, with the most pronounced effects on long-document datasets. Removing sequential edges (w/o Seq) consistently causes the largest degradation on GovReport and arXiv, indicating that local order cues help preserve discourse continuity that is not captured by semantic neighbors alone.
Dropping topic representativeness (w/o Rep) also reduces performance across datasets, demonstrating the importance of maintaining balanced topic coverage.
Removing bridge centrality (w/o Bridge) leads to comparable drops, suggesting that preserving connector sentences is important for retaining cross-section evidence chains.
The cycle cue (w/o Cycle) provides smaller but consistent gains, with the strongest effect on GovReport and arXiv, which aligns with our expectation that cycle participation helps protect local closed-loop reasoning patterns in structured documents.
Finally, disabling redundancy suppression (w/o NMS) notably hurts ROUGE-2 on all datasets, consistent with increased redundancy and less efficient budget usage.

\subsection{Hyperparameter Analysis}
\label{sec:hyper}
We study the impact of the hybrid graph fusion weight $\beta$ controlling the balance between semantic similarity edges ($E_{\text{sem}}$) and sequential edges ($E_{\text{seq}}$). Table~\ref{tab:beta_scan} shows that intermediate fusion is better than either extreme. $\beta=0$ (semantic-only) and $\beta=1$ (sequential-only) both degrade performance, especially on long documents, indicating that cross-topic semantic relations are essential for coverage and relevance while local discourse continuity provides complementary information. The best performance is achieved around $\beta=0.75$ on all datasets, suggesting that a sequentially guided graph with a semantic backbone offers the best balance between local coherence and global topical connectivity.
Further experiments on other parameters are in Appendix~\ref{app:sen}.

\subsection{Token Budget Analysis}

Fig. \ref{fig:cost} presents the quality–budget curves of different methods on four datasets, showing how summarization quality changes as the token budget increases. Across all datasets, performance generally improves as the budget increases, with different increasing rates across methods. Our method shows consistently strong curves across datasets and remains especially competitive under tighter budgets, indicating better budget efficiency. The advantage is more evident on long-form datasets (GovReport, arXiv, PubMed), where preserving global structure and cross-sentence links helps quality keep rising as more budget becomes available, while several baselines exhibit smaller marginal gains beyond mid-range budgets.

\section{Conclusion}
In this paper, we propose a training-free and model-agnostic context compression method that selects a compact set of sentences under a strict token budget while preserving relevance, coverage, and coherence. Our framework builds a hybrid sentence graph that combines semantic similarity with local order, and ranks sentences using an interpretable score that integrates task relevance, topic representativeness, bridge centrality, and cycle coverage. Experiments on four datasets show that the proposed framework is competitive with strong baselines and achieves clear gains on long-document benchmarks. Ablations confirm that the designed components contribute to overall performance.
Our results suggest that simple structural priors can serve as an effective and deployable alternative to training-heavy compressors for long-context LLM applications, and provide a practical foundation for adaptive compression strategies to different tasks, domains, and budget constraints.

\bibliography{ref}
\bibliographystyle{tmlr}

\appendix
\renewcommand{\thetable}{\thesection.\arabic{table}}
\setcounter{table}{0}

\section{Appendix A: Implementation Details}
\label{app:impl}

This appendix provides implementation and engineering details to improve reproducibility and to clarify design choices that are only briefly described in the main text. It does not introduce any new methods or conclusions.

\subsection{Sentence Segmentation and Embedding}
\label{app:sent_split}
Before compression, each document is segmented into a sequence of sentences. We use a standard rule-based sentence splitter that relies on punctuation and capitalization cues, which is stable on formal written text such as news articles and scientific papers. To reduce format noise (e.g., broken line wraps or merged paragraphs that create abnormally long sentences), we remove empty segments and merge consecutive fragments that are extremely short. The same segmentation procedure is applied to all methods for fair comparison.

Each sentence is encoded into a dense vector using OpenAI's \texttt{text-embedding-large} model. The embedding model is used as a frozen pretrained encoder throughout the pipeline, without any task-specific fine-tuning. To control cost and to avoid abnormal long sentences affecting representation quality, we truncate each sentence to at most 512 tokens when computing embeddings. This setting rarely affects natural sentences in our datasets. All sentence embeddings are $L_2$-normalized before similarity computation, consistent with Eq.~\eqref{eq:emb2} in the main text.

\subsection{Hybrid Graph Construction}

We construct the semantic graph using cosine similarity between sentence embeddings, forming a mutual $k$-nearest-neighbor (mutual $k$-NN) graph. To support long documents efficiently, we use approximate nearest neighbor search based on HNSW. An undirected semantic edge is kept between two sentences only if each sentence appears in the other's top-$k$ neighbor list. This mutual selection reduces one-way noisy links and keeps the graph sparse. Unless otherwise stated, we use $k=8$ in the main experiments, while sensitivity to $k$ is analyzed in Appendix~\ref{app:k}.

To complement embedding similarity, which may miss local narrative and discourse continuity, we add sequential edges between adjacent sentences. In all experiments, the sequential window size is set to $\Delta=1$, so each sentence is connected only to its immediate predecessor and successor, while we present results of different choices in Appendix~\ref{app:delta}. Sequential edge weights decay exponentially with sentence distance, as defined in Eq.~\eqref{eq:wseq} in the main text. This design injects a lightweight local order bias without imposing hard structural constraints.

The final hybrid sentence graph is obtained by combining semantic edges and sequential edges with a weighted sum, as in Eq.~\eqref{eq:w}. The coefficients $\alpha$ and $\beta$ control the contributions of semantic similarity and sequential structure, with $\alpha+\beta=1$. Unless otherwise stated, we use $\beta=0.75$ and $\alpha=0.25$. This setting is stable across datasets and document lengths, while sensitivity analysis is reported in Section~\ref{sec:hyper}.

\subsection{Structure-aware Scores Computation}

To estimate latent topical structure and encourage coverage, we cluster sentence embeddings using MiniBatch $k$-means for efficiency. The number of clusters follows the square-root heuristic as in Eq.~\eqref{eq:k}, where $N$ is the number of sentences in the document. This choice balances granularity and stability: too small $K$ can merge distinct content, while too large $K$ may amplify noise. Cluster centroids are used only to compute the representativeness score and are not used in graph construction.

Bridge importance is computed via betweenness centrality on the hybrid graph. Since exact betweenness is expensive on large graphs, we use a sampling-based approximation, where the number of sampled sources scales with $\sqrt{N}$ to balance efficiency and stability. To capture local closed connectivity patterns, we perform bounded cycle enumeration on the hybrid graph to identify small cycles. For computational control, we enumerate at most 200 cycles and use a binary indicator of whether a sentence participates in any enumerated cycle, rather than a continuous cycle score.

\subsection{Budgeted Selection and Redundancy Suppression}

Sentence selection follows a greedy strategy based on the final composite score, ranking sentences as in Eq.~\eqref{eq:score} and selecting them until the token budget is reached. To avoid selecting highly redundant sentences, we apply non-maximum suppression (NMS) using cosine similarity: if a candidate sentence has similarity above $\tau=0.92$ with any selected sentence, it is discarded. This threshold effectively filters near-paraphrases while retaining complementary content, with its sensitivity analysis in Appendix~\ref{app:tau}.

\subsection{Determinism and Reproducibility}

Except for clustering initialization and sampling in approximate betweenness computation, the compression pipeline is deterministic given the same input and hyperparameters. All experiments are run with a fixed random seed (\texttt{seed}=2026). The hyperparameters reported in the main text and appendices are kept the same across all datasets unless explicitly varied in ablation settings.

\subsection{Experimental Setup}

\pb{Datasets.} We evaluate on four widely used summarization benchmarks that cover both news-style inputs and long-form documents. \textbf{CNN/DailyMail} is a news summarization benchmark pairing articles with multi-sentence highlights, and it is widely used to evaluate short-to-medium document summarization~\cite{hermann2015teaching}. \textbf{GovReport} is a long-document summarization dataset built from government reports, designed to test model robustness when salient evidence is distributed across many sections~\cite{huang2021efficient}. \textbf{arXiv} is a scientific-paper summarization dataset where the goal is to generate an abstract-level summary from a full paper, emphasizing long-range structure and section-level coherence~\cite{cohan-etal-2018-discourse}. \textbf{PubMed} is a biomedical-paper summarization dataset with the same setup as arXiv, stressing factual precision and coverage over long inputs~\cite{cohan-etal-2018-discourse}.

\pb{Baselines.} We include both extractive and abstractive systems to cover different compression designs under budget constraints. \textbf{TextRank} is an unsupervised extractive method that builds a sentence similarity graph and ranks sentences using graph centrality~\cite{mihalcea2004textrank}. \textbf{LEAD-3} is a strong positional heuristic that selects the first three sentences, and it remains a competitive reference point on news-style data. \textbf{BERTSum} is a supervised extractive framework that fine-tunes pretrained encoders to score and select salient sentences~\cite{liu2019bertsum}. \textbf{MatchSum} formulates extractive summarization as semantic matching between a candidate sentence set and the source document to better align subset selection with document meaning~\cite{zhong2020matchsum}. \textbf{MemSum} is a multi-step extractive approach that models selection history to reduce redundancy and improve coverage, which is useful for longer documents~\cite{gu2022memsum}. \textbf{RankSum} is an unsupervised extractive method that fuses multiple sentence-ranking signals (e.g., topic, semantic, keyword, position) to derive a final salience ranking~\cite{joshi2022ranksum}. 

\textbf{BART} is a pretrained seq2seq denoising model commonly used as an abstractive summarization backbone~\cite{lewis2020bart}. \textbf{PEGASUS} is a summarization-oriented pretraining method that uses gap-sentence generation to improve abstractive summarization quality~\cite{zhang2020pegasus}. \textbf{BIGBIRD} extends Transformers to longer inputs via sparse attention, enabling long-context summarization with reduced attention cost~\cite{zaheer2020big}. \textbf{DANCER} targets long-document summarization by leveraging discourse-aware decomposition and recombination to improve scalability~\cite{gidiotis2020dancer}. \textbf{HiGen} is a planning-style long-document summarization method that uses highlight guidance (sentence-level plans) to steer generation toward salient content~\cite{du2025higen}. \textbf{SigExt} is a prompt-based summarization approach that extracts salient signals (e.g., keyphrases/highlights) to improve content control and faithfulness during generation~\cite{xu2024sigext}.

\pb{Metrics.} \textbf{ROUGE-1/2/L} measures n-gram and longest-common-subsequence overlap between system summaries and references and is a standard lexical metric for summarization~\cite{lin2004rouge}. \textbf{BERTScore} evaluates semantic similarity using contextual embeddings, providing a complement to surface-overlap metrics when paraphrasing is common~\cite{zhang2019bertscore}. \textbf{QAFactEval} is a QA-based factual consistency metric that checks whether reference-supported facts are preserved in the generated summary, which is especially relevant for long, information-dense documents~\cite{fabbri2022qafacteval}.

\section{Appendix B: Hyperparameter Sensitivity and Robustness Analysis}
\label{app:sen}

This appendix reports systematic sensitivity and robustness analyses of key hyperparameters in our framework. Building on the hyperparameter analysis in Section \ref{sec:hyper} in the main text, the experiments here evaluate stability over continuous or graded parameter settings, including the semantic neighbor size $k$, the redundancy suppression threshold $\tau$, the sequential window size $\Delta$, and the scoring weights. Unless otherwise stated, all experiments are conducted on GovReport, and we report ROUGE-L as the primary metric. We also observe consistent trends with ROUGE-1/2, BERTScore, and QAFactEval.

\subsection{Sensitivity to Semantic Neighbor Size $k$}
\label{app:k}
In the semantic similarity graph, each sentence node connects to other sentences only through mutual $k$-nearest-neighbor relations. A small $k$ can make the graph too sparse and fragment connectivity, while a large $k$ may introduce redundant edges and reduce sparsity. 
As shown in Table \ref{tab:hyper_sens_all}, performance varies only mildly in the range $k\in[6,12]$, suggesting that our method is not highly sensitive to the choice of $k$ within a reasonable band. The default setting $k=8$ lies in this stable region and achieves the best ROUGE-L in our sweep.

\subsection{Sensitivity to NMS Threshold $\tau$}
\label{app:tau}
Non-maximum suppression (NMS) controls redundancy filtering through the cosine similarity threshold $\tau$. Table~\ref{tab:hyper_sens_all} shows that performance is stable for $\tau\in[0.90,0.95]$. A smaller threshold can over-filter and reduce coverage, while a larger threshold allows more redundancy. The default $\tau=0.92$ provides a good balance between coverage and redundancy, consistent with the ``w/o NMS'' ablation in the main text.

\subsection{Sensitivity to Sequential Window Size $\Delta$}
\label{app:delta}
We use sequential edges to capture local discourse continuity that may not be reflected by semantic neighbors alone. Table~\ref{tab:hyper_sens_all} shows that adding sequential edges ($\Delta=1$) results in a clear improvement over removing them ($\Delta=0$). Increasing the window beyond $\Delta=1$ slightly reduces ROUGE-L, suggesting that a minimal local-order bias is sufficient and larger windows may introduce redundant connections.

\begin{table}[t]
\centering
\small
\setlength{\tabcolsep}{6pt}
\renewcommand{\arraystretch}{1.15}
\begin{tabular}{lcccc}
\toprule
$k$ & $k{=}4$ & $k{=}6$ & $k{=}8$ & $k{=}12$ \\
ROUGE-L & 51.24 & 51.78 & \textbf{52.01} & 51.85 \\
\midrule
$\tau$ & $\tau{=}0.88$ & $\tau{=}0.90$ & $\tau{=}0.92$ & $\tau{=}0.95$ \\
ROUGE-L & 51.46 & 51.78 & \textbf{52.01} & 51.72 \\
\midrule
$\Delta$ & $\Delta{=}0$ & $\Delta{=}1$ & $\Delta{=}2$ & $\Delta{=}3$ \\
ROUGE-L & 49.28 & \textbf{52.01} & 51.74 & 51.62 \\
\bottomrule
\end{tabular}
\caption{Sensitivity analysis on GovReport (ROUGE-L) for the semantic neighbor size $k$, the NMS threshold $\tau$, and the sequential window size $\Delta$.}
\label{tab:hyper_sens_all}
\end{table}

\subsection{Sensitivity to Scoring Weights}

We further study the robustness of the scoring function by jointly scanning the four scoring weights $\{\lambda_{\text{task}}, \lambda_{\text{rep}}, \lambda_{\text{bridge}}, \lambda_{\text{cycle}} \}$. All weights satisfy $\lambda_{\text{task}} + \lambda_{\text{rep}} + \lambda_{\text{bridge}} + \lambda_{\text{cycle}} = 1$, and each $\lambda \geq0$. The goal of this scan is not to search for an optimal setting, but to verify that performance remains stable under reasonable weight allocations.

\pb{Setup.} We perform a coarse sweep over the task weight $\lambda_{\text{task}} = \{0.35, 0.45, 0.55, 0.65\}$. Given $\lambda_{\text{task}}$, the remaining structural budget is $1-\lambda_{\text{task}}$. We consider three representative allocation patterns over $(\lambda_{\text{rep}}, \lambda_{\text{bridge}}, \lambda_{\text{cycle}})$:
\[
\begin{aligned}
    \mbox{{Rep-heavy}: } &(0.60, 0.30, 0.10) \cdot (1-\lambda_{\text{task}}), \\
    \mbox{Balanced: } &(0.50, 0.40, 0.10) \cdot (1-\lambda_{\text{task}}), \\
    \mbox{Bridge-heavy: } &(0.40, 0.50, 0.10) \cdot (1-\lambda_{\text{task}}).
\end{aligned}
\]
In addition, when $\lambda_{\text{task}}=0.45$, we include a \textbf{Balanced} setting without the cycle term ($\lambda_{\text{cycle}}=0$) to directly align with the \emph{w/o Cycle} ablation in the main text.

\begin{table}[t]
\centering
\small
\setlength{\tabcolsep}{3pt}
\renewcommand{\arraystretch}{1.1}
\begin{tabular}{c l c c c c}
\toprule
$\lambda_{\text{task}}$ & Setting & $\lambda_{\text{rep}}$ & $\lambda_{\text{bridge}}$ & $\lambda_{\text{cycle}}$ & ROUGE-L \\
\midrule
0.35 & Rep-heavy & 0.39 & 0.20 & 0.06 & 51.74 \\
0.35 & Balanced & 0.33 & 0.26 & 0.06 & 51.80 \\
0.35 & Bridge-heavy& 0.26 & 0.33 & 0.06 & 51.68 \\
\midrule
0.45 & Balanced (Full) & 0.28 & 0.23 & 0.05 & \textbf{52.01} \\
0.45 & Rep-heavy & 0.33 & 0.17 & 0.05 & 51.86 \\
0.45 & Bridge-heavy & 0.22 & 0.28 & 0.05 & 51.78 \\
0.45 & Balanced (w/o Cycle) & 0.31 & 0.24 & 0.00 & 51.17 \\
\midrule
0.55 & Rep-heavy & 0.27 & 0.14 & 0.04 & 51.62 \\
0.55 & Balanced & 0.23 & 0.18 & 0.04 & 51.69 \\
0.55 & Bridge-heavy& 0.18 & 0.23 & 0.04 & 51.55 \\
\midrule
0.65 & Rep-heavy & 0.21 & 0.11 & 0.03 & 51.18 \\
0.65 & Balanced & 0.18 & 0.14 & 0.03 & 51.30 \\
0.65 & Bridge-heavy& 0.14 & 0.18 & 0.03 & 51.06 \\
\bottomrule
\end{tabular}
\caption{Joint scan of scoring weights on GovReport (ROUGE-L).}
\label{tab:weight_joint_scan}
\end{table}

\pb{Analysis and consistency with ablations.}
The results in Table~\ref{tab:weight_joint_scan} indicate a stable operating region for the task weight: performance is relatively consistent when $\lambda_{\text{task}} \in [0.35, 0.55]$, while placing too much weight on the task term (e.g., $\lambda_{\text{task}}=0.65$) reduces the contribution of structural signals and leads to clearer degradation. For a fixed $\lambda_{\text{task}}$, different allocations among representativeness and bridge centrality yield only modest differences, suggesting that the best region is not an isolated point but a smooth neighborhood. Finally, removing the cycle term under the same base setting (Balanced, $\lambda_{\text{task}}=0.45$) produces a noticeable drop, matching the qualitative pattern in the \emph{w/o Cycle} ablation. Overall, these findings suggest that the gains observed in the main experiments do not rely on delicate weight tuning, but rather on the complementary interaction between task relevance and multiple structural priors.

\end{document}